\documentclass[11pt,a4paper]{article}
\usepackage[hyperref]{emnlp2020}
\usepackage{times}
\usepackage{latexsym}

\usepackage{graphicx}
\newcommand{\xdownarrow}[1]{%
  {\left\downarrow\vbox to #1{}\right.\kern-\nulldelimiterspace}
}
\usepackage[many]{tcolorbox}

\usepackage{microtype}

\usepackage{amsmath}

\usepackage{array}

\newcolumntype{M}[1]{>{\centering\let\newline\\\arraybackslash\hspace{0pt}}m{#1}}
\newcolumntype{L}[1]{>{\centering\let\newline\\\arraybackslash\hspace{0pt}}m{#1}}
\newcolumntype{C}[1]{>{\centering\let\newline\\\arraybackslash\hspace{0pt}}m{#1}}
\newcolumntype{R}[1]{>{\centering\let\newline\\\arraybackslash\hspace{0pt}}m{#1}}
\newcolumntype{P}[1]{>{\centering\let\newline\\\arraybackslash\hspace{0pt}}m{#1}}
\newcolumntype{Q}[1]{>{\centering\let\newline\\\arraybackslash\hspace{0pt}}m{#1}}
\newcolumntype{S}[1]{>{\centering\let\newline\\\arraybackslash\hspace{0pt}}m{#1}}

\aclfinalcopy 

\definecolor{KDpurple}{rgb}{0.6,0.18,0.64}

\title{Resolving Intent Ambiguities by Retrieving Discriminative Clarifying Questions}

\author{Kaustubh D. Dhole \\
  Amelia Science \\
  R\&D, IPsoft \\
  New York, NY 10004 \\
  \texttt{\textcolor{darkblue}{kdhole@ipsoft.com}}}

\date{}

\begin{document}
\maketitle
\begin{abstract}
Task oriented Dialogue Systems generally employ intent detection systems in order to map user queries to a set of pre-defined intents. However, user queries appearing in natural language can be easily ambiguous and hence such a direct mapping might not be straightforward harming intent detection and eventually the overall performance of a dialogue system. Moreover, acquiring domain-specific clarification questions is costly. In order to disambiguate queries which are ambiguous between two intents, we propose a novel method of generating discriminative questions using a simple rule based system which can take advantage of any question generation system without requiring annotated data of clarification questions. Our approach aims at discrimination between two intents but can be easily extended to clarification over multiple intents. Seeking clarification from the user to classify user intents not only helps understand the user intent effectively, but also reduces the roboticity of the conversation and makes the interaction considerably natural.
\end{abstract}

\section{Introduction}

Task oriented dialogue systems aim at extracting semantic information from natural language queries in order to decipher user's intents. Such systems play a vital role in commercial applications like personal assistants (e.g. Google Home, Alexa,  Siri,  etc.) for a variety of domain specific tasks like flight-booking, call routing, restaurant booking and so on which typically model a dedicated Natural Language  Understanding  (NLU)  component  that  performs  inference for downstream tasks like domain classification, intent detection and slot filling. 

A major driving component of NLU is Intent Detection which operates over users' queries. However, users' queries are generally ambiguous and underspecified. Eg. in a banking domain, given two pre-defined intents, ``opening\_a\_savings\_account'' and ``opening\_a\_checking\_account'', even a simple user query like ``I want to open an account'' does not directly map to either of the two intents and requires disambiguation. Managing this would require creating a separate intent representing ``opening an account'' but that would mean creating the corresponding task workflows, acquiring extra training data to incorporate the new intent and retraining intent and possibly other subsequent classifiers.\footnote{While this also depends on the design of the dialog system, we assume the pipelined approach of classifying the domain first followed by the intent.} 

\begin{figure}
  \includegraphics[width=\linewidth]{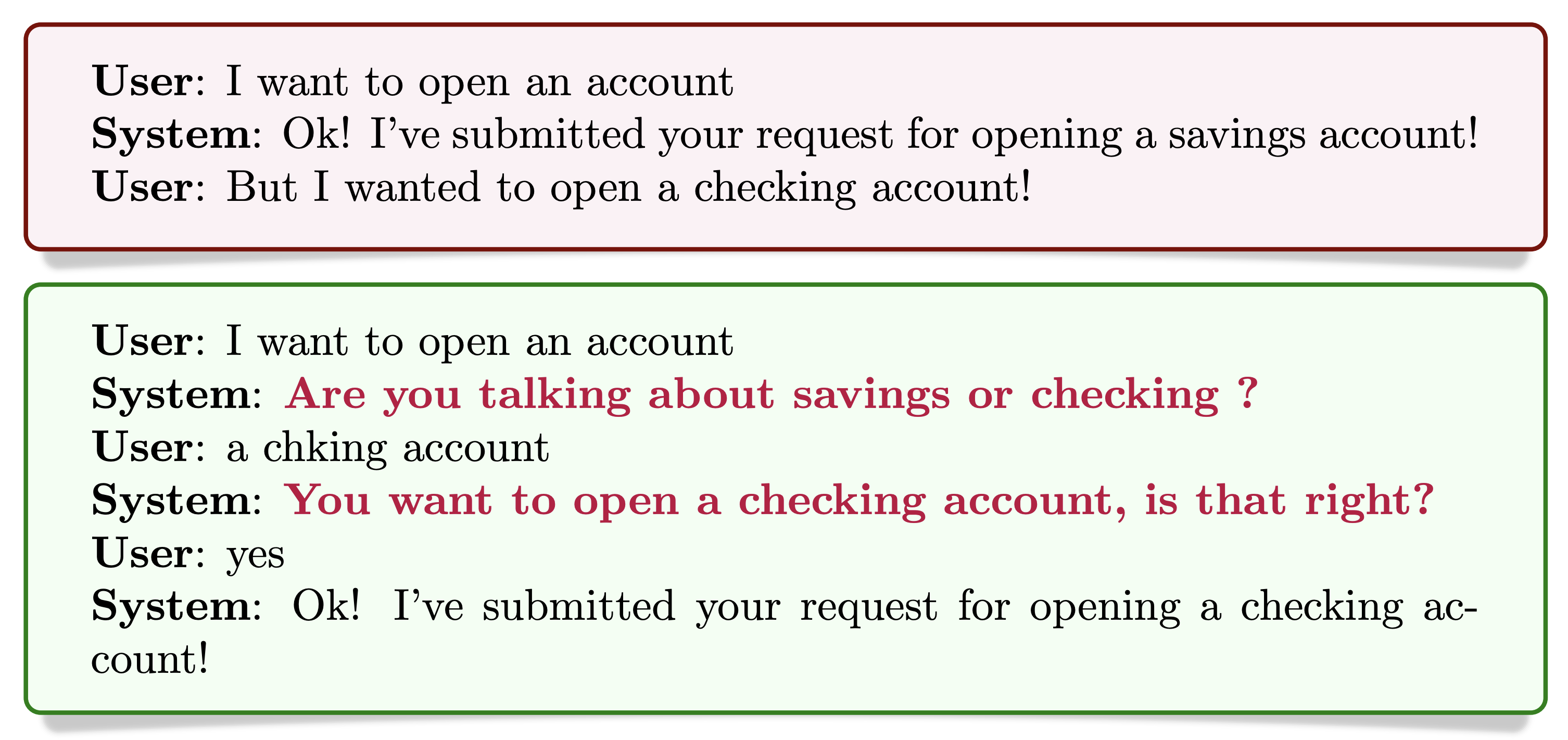}
  \caption{In the first conversation, the system suffers due to unavailability of a separate intent and hence misunderstands the user's intent. In the second conversation, the system generates two clarifying questions in order to disambiguate and clarify the user's intent successfully.}
  \label{fig:cqa_pass}
\end{figure}
In this paper, we explore this problem specifically in the more pragmatic task oriented dialog setting to improve intent classification by incorporating a limited form of unsupervised interaction as shown in Figure~\ref{fig:cqa_pass}. In order to disambiguate between two intents, given an ambiguous natural language query, we describe a simple rule-based system to generate discriminative questions using an existing question generator and a sentence similarity model. Generating discriminative questions has significant advantages over a one-to-one utterance-to-intent classification: (i) It improves the overall accuracy of classifying the user's intent since it boils down the role of non-deterministic classifiers from a top-1 to an easier top-k classification problem permitting the classifiers a little slack in performance by acquiring clarification from the end-user herself. (ii) Rather than relying on a single user input, the communication with the dialog system becomes highly interactive.

\section{Related Work}
Clarification requests were studied in dialogue extensively by~\cite{purver2003means,purver2003answering, purver2004theory, healey2003experimenting} who also established a taxonomy of the various types of clarification.~\citet{Coden2015DidYM} discussed challenges involved in disambiguating entities via clarification. With the rise of conversational systems, there has been enormous interest in generating clarifying questions and datasets recently.~\citet{DBLP:conf/emnlp/XuWTDYZZS19} constructed a clarification dataset to address ambiguity arising in knowledge-based question answering.~\citet{aliannejadi2019asking} proposed a clarification dataset to improve open-domain information-seeking conversations.~\citet{kumar-black-2020-clarq} built a clarification dataset by sampling comments from StackExchange posts.~\citet{DBLP:journals/corr/abs-1904-02281, cao-etal-2019-controlling, Zamani2020GeneratingCF} have attempted to use neural models to train over (context, question) pairs to generate clarifying questions.~\citet{DBLP:journals/corr/abs-1904-02281} proposed an RL based model for generating a clarifying question in order to identify missing information in product descriptions.~\citet{cao-etal-2019-controlling} described an interesting approach feeding expected question specificity along-with the context to generate specific as well as generic clarifying questions. However, most of these models still require large amounts of training data with Wizard-of-Oz style dialog annotations.~\citet{yu2019interactive} attempt to generate binary and multiple choice questions and show the benefit of incorporating interaction for determining user's intent.

~\citet{xu2020distinguish} use a graph neural network and a novel attention mechanism to capture the discriminative attributes of confusing law articles.Emphasizing that ambiguity is a function of the user query and the evidence provided by a very large text corpus, ~\citet{min2020ambigqa} introduce an interesting dataset and an associated task to generate disambiguating rewrites of an original open-domain question.~\citet{li2017learning} explore an effective method for generating discriminating questions to disambiguate pairs of images.

Our approach is close to~\citet{yu2019interactive, Zamani2020GeneratingCF}. In contrast to~\citet{yu2019interactive} where questions and answer choices are manually generated, we seek to automate this by using a simple TF-IDF approach to generate potential answer choices to disambiguate to a particular intent. Additionally, instead of collecting domain specific discriminative questions which are harder to obtain, we show how we can generate discriminative questions by using only a sentence-level question generator and a discriminative similarity measure. Besides, our approach is simpler to incorporate in production systems with small amounts of training data for intents. Keeping this interaction component partitioned from the one-to-one intent classifier also eases its incorporation into dialogue systems with pre-deployed intent classifiers. 

\section{Model}
Given an ambiguous utterance, our goal is to generate a discriminative question to obtain clarification between two highly probable intents. 
\begin{itemize}
\itemsep0em 
\item First, we train an intent classifier and classify the incoming utterance into one of several intents
\item Using a pre-trained question generation system, we generate question answer pairs and select the question with the highest potential to discriminate
\item If the question does not have high \textit{discriminative similarity}, we generate a template based question from the intent's pre-computed discriminative attributes
\item Finally, we classify the user's subsequent response to the discriminative question into either of the two intents or none of them.
We describe each of the steps in detail in the following subsections.
\end{itemize}
\subsection{Intent Classification and Ambiguity Detection}
 Given a set of utterances in the form of user sentences $x_1, x_2 ... x_n$ with their annotated intents $y_1, y_2 ... y_n$ where $y_i \in {1 ... m} $, we train a sentence classifier in order to create an intent classifier. A softmax layer is used to assign probabilities to each intent $p_1, p_2 ... p_n$.
 
 At runtime for a given user query $q$, we execute the intent classifier to get the probability scores. 
 
 If the softmax scores of the highest intent $j$ namely $p_j$ is lesser than a pre-determined confidence threshold $t_1$, we consider the query as ambiguous in itself.
 
 If the softmax scores of the two highest intents $j, k $ namely $p_j, p_k$ is within a pre-determined threshold, we consider the query as ambiguous between intents $j$ and $k$ i.e. if $p_j - p_k < t_2$ where $t_2 \in [0,1]$ is the two-intent ambiguity threshold.
 
\subsection{Discriminative Question Selection}
\label{ssec:dqs}
In order to generate a discriminative question, we use an existing question generation system and the annotated utterances used for training the intent classifier itself. We collect all the training utterances of the top two ambiguous intents:
\begin{gather*}
J = \{x_i \forall i | y_i=j\} \\
K = \{x_i \forall i | y_i=k\}
\end{gather*}
For all the utterances in $J$ and $K$, we generate question answer pairs using SynQG~\cite{dhole-manning-2020-syn} and accumulate them in the following two sets of question answer pairs respectively from which we select one question-answer pair from each set in order to further compute our representative discriminative question. 
\begin{gather*}
Q_J, Q_K
\end{gather*}
We are interested to select a question-answer pair from each of the above two sets whose question can serve as a potential discriminative question. We attempt to identify one question-answer pair from each set $(q_J^{*}, a_J^{*}) \in  Q_J$ and $(q_K^{*}, a_K^{*}) \in  Q_K$ using the following discriminatory conditions. 

\begin{gather*}
\forall j \in  |Q_J|, \forall k \in  |Q_K| :\\
\hspace{60pt} s_{j,k} = score(q, q_j, a_j, q_k, a_k)\\
{j^{*}, k^{*}} = max(s_{j,k} \forall (j,k))
\end{gather*}

where the discriminative score is defined as follows:
\begin{gather*}
    score(q, q_j, a_j, q_k, a_k) = sim(q_j, q_k)\\
    - sim(a_j, a_k) \\
    + 0.5 (sim(q, q_j) + sim(q, q_k) )
\end{gather*}

We hypothesize that an ideal discriminative question would be such that its corresponding answers for each of the intents would have to be not only different and discriminative, but the answers should exclusively be present in each of the two intents. Hence we would expect $a_J^{*}$ and $a_K^{*}$ to be highly dissimilar: $- sim(a_j, a_k)$\footnote{This also depends on the choice of the similarity function like retro-fitting vectors~\cite{faruqui-etal-2015-retrofitting} might be a better choice than GloVe~\cite{pennington-etal-2014-glove} or Word2Vec~\cite{mikolov2013distributed} when answers are common nouns or adjectives.}

Additionally, both $q_J^*$ and $q_K^*$ should be neutral to both the intents and highly similar to each other since we want a question to be identical enough to trigger both the intents: $sim(q_j, q_k)$

We draw a parallel here with~\citet{li2017learning}'s task of generating a discriminative question from a pair of ambiguous images by seeking to identify discriminative regions, choosing pairs with high contrast, high visual dissimilarity and high question similarity.

However, two intents might have multiple sources of ambiguity than provided by their definitions. We try to figure out the specific source by looking at the user utterances used within the two intents and the user query. Consider the following extreme case wherein the first question in both the given pairs belongs to a common intent and the second question also belongs to another common intent : The following pair qualifies as being rightly disambiguating:

\textit{(What is the type of account?, savings)} 

\textit{(What is the account type?, checking)} \\
as well as this pair:

\textit{(What would you like to do?, open a savings account)}

\textit{(What do you want to do?, open a checking account)}

For a user query $q=$ ``I want to open an account'', questions belonging to the first pair can serve as discriminatory questions but not from the second pair. However, for a user query like $q=$ ``I would like do do this'', only questions from the second pair would be useful. Hence, we attempt to further re-rank the questions by the similarity with the user query: $+ 0.5 (sim(q, q_j) + sim(q, q_k) )$. 

For each of the above similarity computations, we use encodings from~\citet{cer2018universal} and perform a cosine similarity.

Moreover, since user queries' grammatical style is meant to be in a form suitable to communicate facing the agent, these questions can't be presented back to the user directly. And hence, we perform a simple set of substitutions to perform the conversion:

\begin{itemize}
\item ``I VERB'' to ``you want/need to VERB'' when the main verb is not need/want
\item ``you, your, etc.'' to ``me, mine, etc.''
\item If the main verbs of $q_J^*$ and $q_K^*$ are the same and both have the same direct object, with different modifiers, then use the hypernym of the modifiers in a type question or provide both the modifiers to the user
\item If the main verbs of $q_J^*$ and $q_K^*$ are the same and have no children, then use it and display both the answers too to the user $a_J^{*}$ and $a_K^{*}$.
\item If the main verbs are not the same, check if a hypernym exists and use it instead of the main verb
\begin{gather*}
q_{d} = combine(q_J^*, a_J^*, q_K^*, a_K^*)
\end{gather*}
\item In the end, we perform a back-translation using the same translators as~\citet{dhole-manning-2020-syn}
\end{itemize}

\subsection{Template Based Clarification Question}
\label{ssec:tbcq}
Not all pairs of intents would generate highly discriminative questions with the above approach because of either lack of enough training examples or the user presenting novel sources of ambiguity unseen in the training data or possessing verbs which are completely different. In order to deal with such cases, we use a handful generalized templates like ``Are you talking about $DP_j$ or $DP_k$?'' where $DP_i$ is a discriminative phrase of intent $i$. We pre-compute TF-IDF n-grams from the training data itself and use them as discriminative phrases. Since they serve as candidate answers which users can select from, it helps steer the conversation towards a structured path which has a guaranteed back-end workflow.

\subsection{Ambiguity Resolution}
Our final step is to decipher the user's response $r$ to the clarifying question and classify it into either of the two intents or none of them. We do this by computing sentence encodings from~\citet{cer2018universal} and then perform a cosine similarity of $r$ with $a_J^*$, $a_K^*$ and $N_o$ each to identify which of the three options is the closest, where $N_o$ is a set of commonly spoken keywords like ``none'', ``none of them'' etc.
\begin{table*}
\centering
\small
\begin{tabular}{M{3cm} | L{2cm} | C{2cm}}
\hline
\textbf{Classifier} &
\textbf{F1-score (Top-1)} &
\textbf{F1-score (Top-2)}\\
\hline
& & \\
BiLSTM                  & 89.23 & 93.09 \\
Linear SVM              & 82.58 & 85.86 \\
\hline
\end{tabular}
\caption{Intent classification Performance}
\label{top-2-table}
\end{table*}
\section{Experiments and Results}
We use a commercial data-set of user queries belonging to an IT service desk domain. This dataset has 8,700 (train) + 3800 unambiguous (test) + 1068 ambiguous (test) user queries in the form of sentences annotated with 80 intents. Each of the sentences has been generated via crowd-sourcing over Amazon Mechanical Turk. To create the unambiguous train and test sets, each worker was provided an intent description and a few seed example utterances as references. We also created an ambiguous set, which was only used for testing, for which workers were provided the intent descriptions and the seed examples of two intents and were asked to come up with an utterance which would either be a generic or an abstract version of both the intents or could single-handedly serve as a representative for both~\footnote{On manual analysis, 38 out of 100 randomly chosen examples were found to represent both the expected user intentions explicitly rather than a common abstracted representation. This is understandable for intent pairs which hardly have anything in common at all ~\textit{- Hey, can I get someone to help me~\textbf{archive emails},and also I ~\textbf{want to start excel inside a VM}.}}.
We train two sentence classifiers using a BiLSTM and a linear SVM~\cite{fan2008liblinear}. 

It is imperative to find out to what extent would asking a clarifying question benefit quantitatively. In the case of this dataset, we get a potential bandwidth to increase the F1-score by around 4\% for a BiLSTM and around 3\% for a linear SVM. (Table~\ref{top-2-table})

\begin{figure}
  \includegraphics[width=\linewidth]{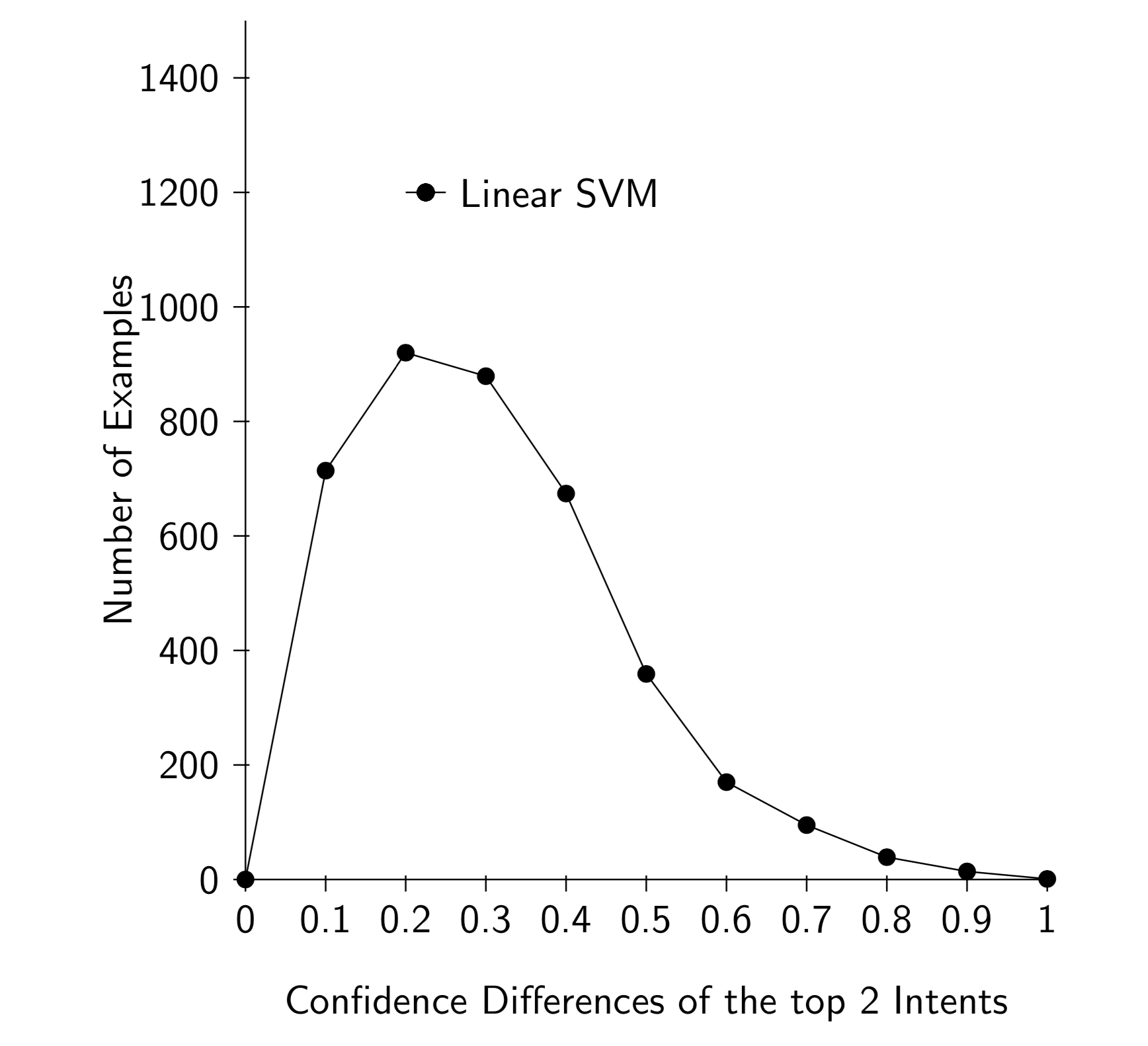}
  \caption{Due to low separation between the top-2 predicted intent classes, most of the test set examples (single class examples) have the first and second highest intent predictions extremely close resulting in false ambiguities.}
  \label{fig:graph1}
\end{figure}

\begin{figure}
  \includegraphics[width=\linewidth]{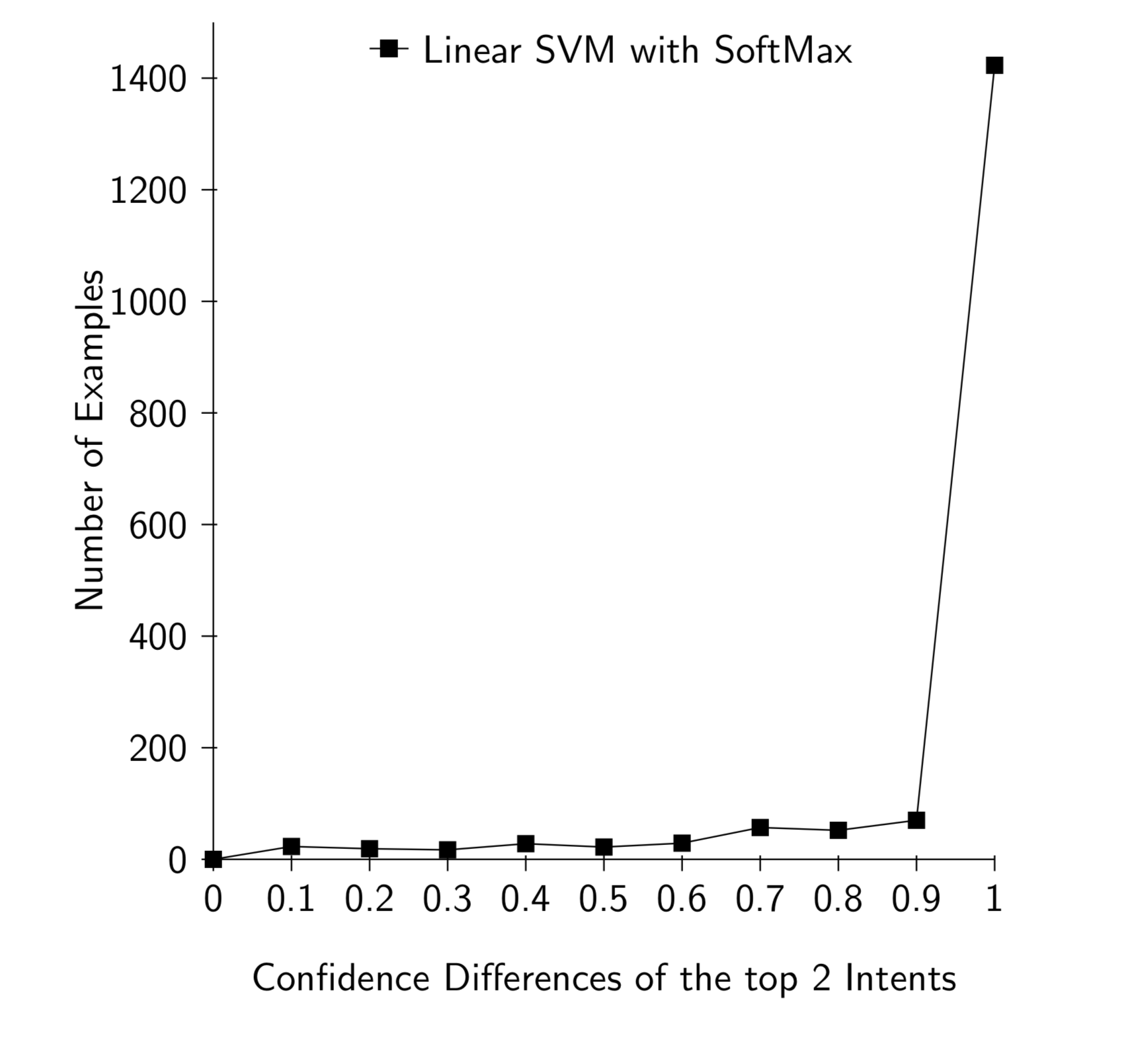}
  \caption{After softmax calibration, the confidence separation increases further apart and almost all of the examples are correctly pushed in the unambiguous zone on the right.}
  \label{fig:graph2}
\end{figure}

It is easier to define the ambiguity threshold $t_2$ by looking at the confidence scores of the predictions. However, we also need to ensure that such thresholds avoid false positives in assessing ambiguity. We notice that for a linear SVM, a large number of predictions are false positives. This is because the difference between the top-2 intents is reasonably close (Figure~\ref{fig:graph1}). We notice that after calibrating~\cite{guo2017calibration} these scores with a softmax layer, the predictions become highly confident as the difference between the top-2 intents' confidences increases as shown in Figure~\ref{fig:graph2}.

\subsection{Performance of Ambiguity Detection}
We compare the predictions of both the trained models on the ambiguous test set. We mark a prediction as correct if the top-2 predicted intents have close scores and both of them match the two expected intents~\footnote{We do not put a check on the order of the two intents.}. The threshold based parameter $t_2=0.3$ is able to identify the top-2 intents in 839 and 709 out of 1068 cases for a linear SVM and a BiLSTM respectively. 
\begin{figure}
  \includegraphics[width=\linewidth]{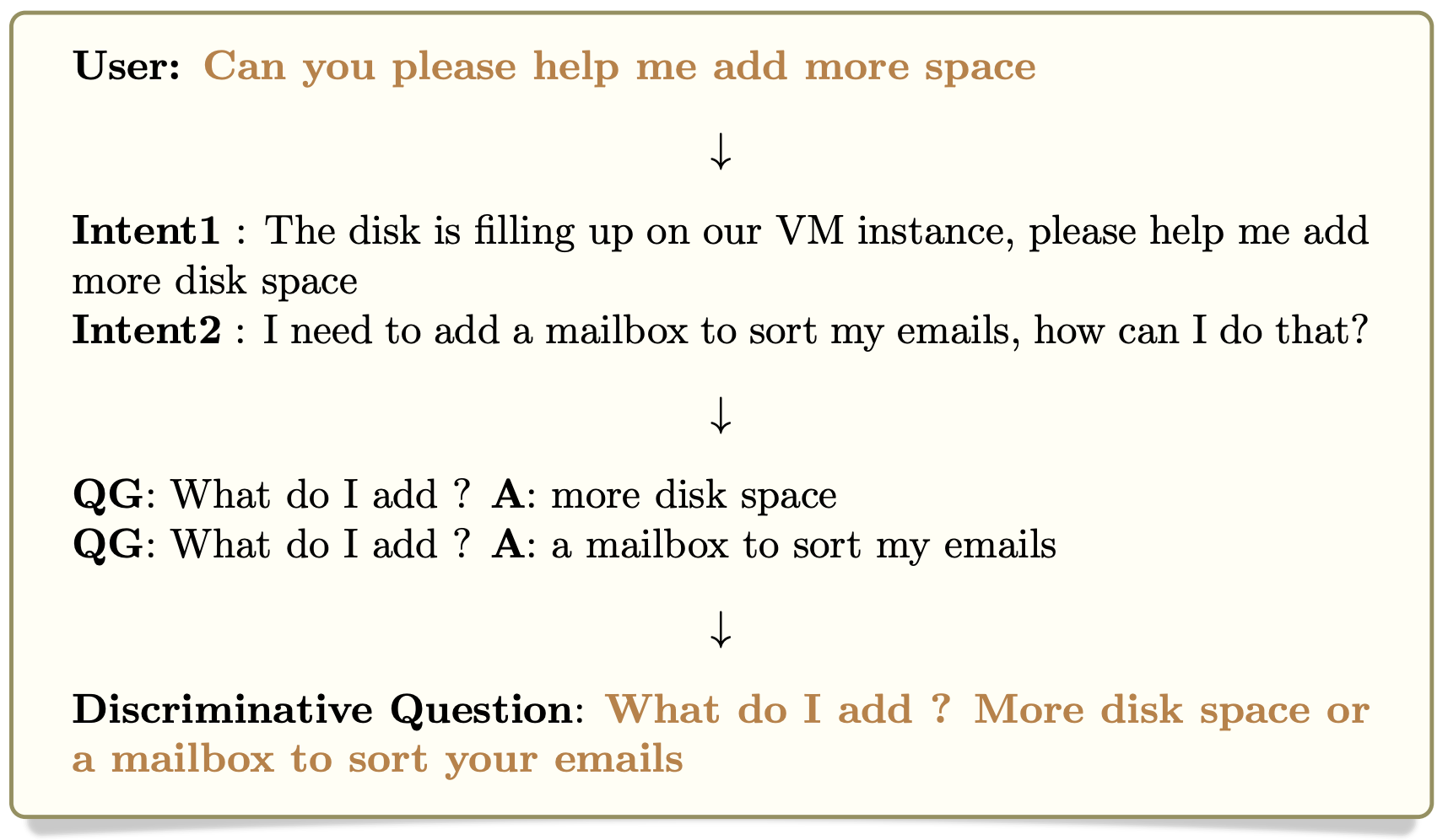}
  \caption{Here, the two representative utterances as well as the user utterance possess the same verb ``add''. The user query is ambiguous and needs clarity as to what needs to be added.}
  \label{fig:d1}
\end{figure}
\begin{figure}
  \includegraphics[width=\linewidth]{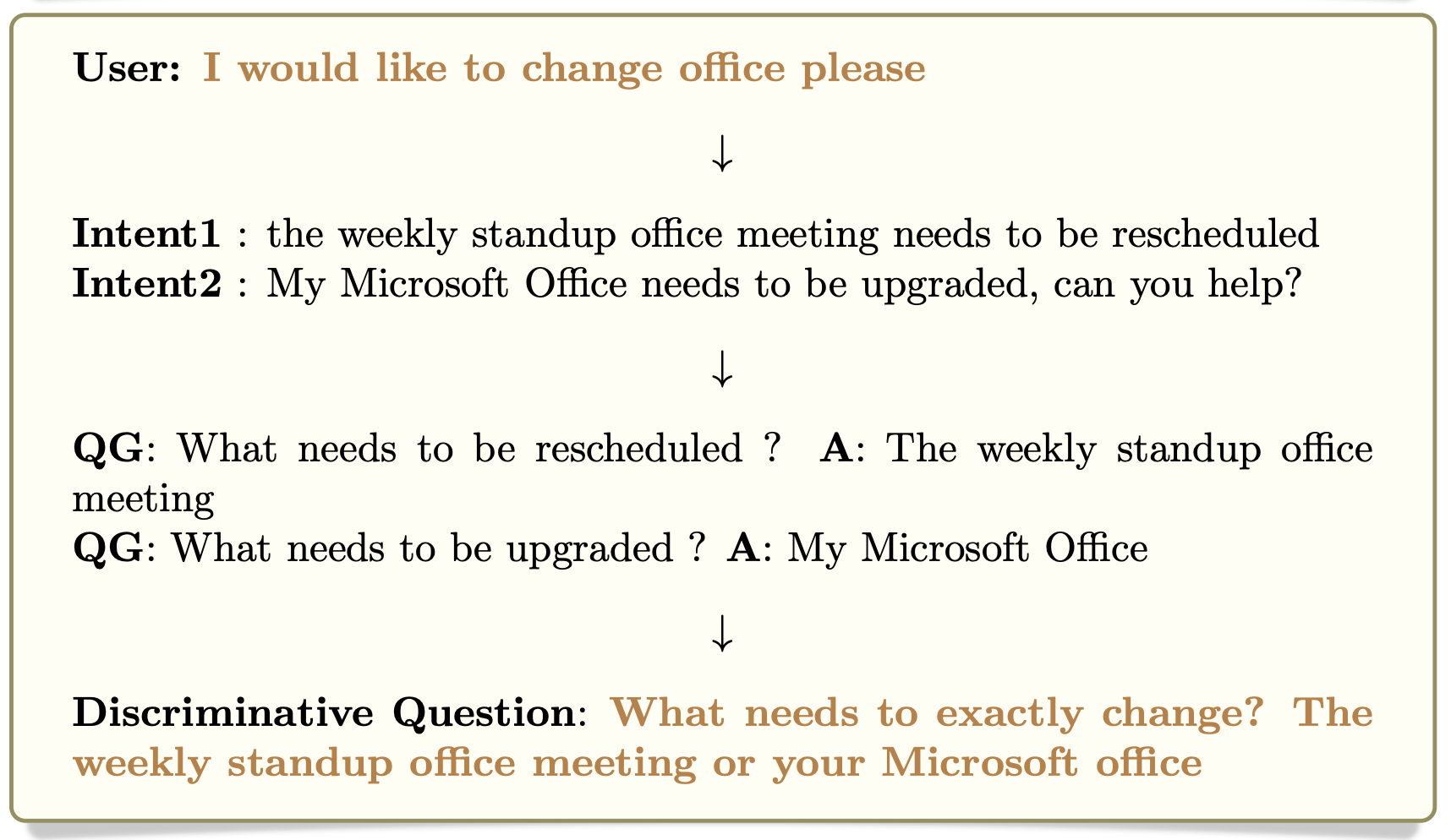}
  \caption{Here, the user phrase ``office'' needs to be disambiguated. Both of the representative utterances picked here refer to the action of ``change' which is a hypernym.}
  \label{fig:d2}
\end{figure}
\begin{figure}
  \includegraphics[width=\linewidth]{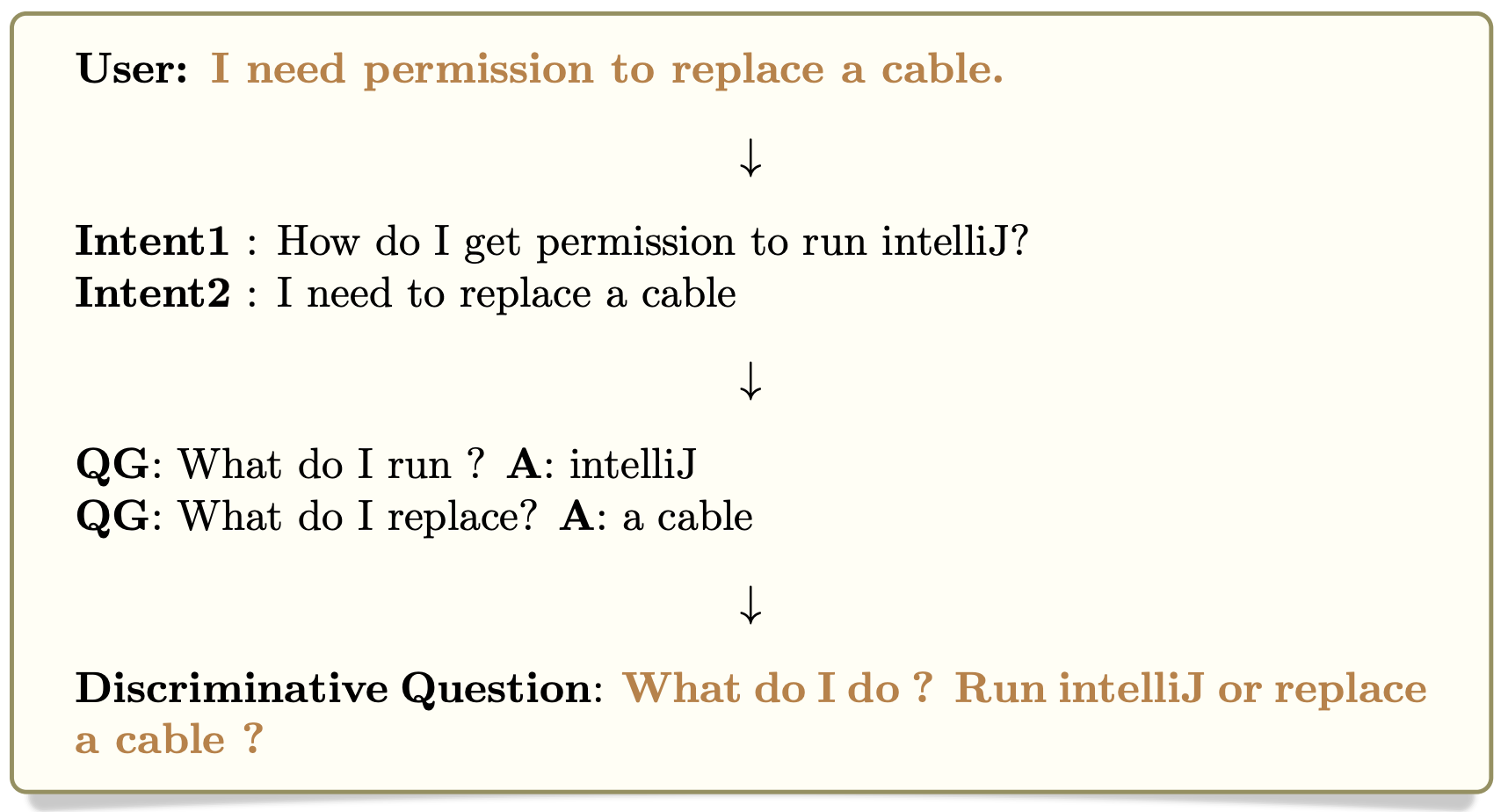}
  \caption{In this case, the verbs are completely different and hence a generic verb ``do'' is used alongwith the two answer options.}
  \label{fig:d3}
\end{figure}

\subsection{Performance of Discriminative Questions}
In order to evaluate the performance of discriminative questions, we select 100 examples from the ambiguous test set which have been detected as ambiguous by both the classifiers and generate a discriminative question using the procedure described in section~\ref{ssec:dqs}. We request MTurk raters to evaluate the grammaticallity and relevance of the generated questions by utilizing the settings of SynQG. Instead of a single fact, we ask raters to look at the corresponding 2 source utterances while gauging relevance. The average grammaticality and relevance are found to be 3.84 and 4.17 respectively on a 5-point Likert scale close to that of SynQG.
We also find that questions generated using only the QG approach (and not the template based approach) depict a poor coverage of 34\% due to missing common verbs or lack of enough generated question pairs. 
We show three examples in figures~\ref{fig:d1} to~\ref{fig:d3}.

\section{Discussion}
We seek to improve intent classification and enhance user interaction by detecting the presence of ambiguity and making the user answer discriminative questions by generating questions from an existing sentence to question generator. Such a rule-based approach which is segregated from the intent identification logic is easy to deploy onto conversation systems with pre-existing intent classifiers. However, the coverage of the discriminative similarity approach is still low and holds tremendous scope for improvement. Nevertheless, for conversation systems, such an approach can still be used to reduce manual effort for pre-generating discriminative questions by removing the dependency on the runtime user query $q$ from the discriminative similarity measure equation. Also, correctly identifying discriminative attributes as a first step will still be a key to generate strong discriminative questions as validated in the visual counterpart~\cite{li2017learning}. While our approach identifies such attributes in the surface forms of user utterances, scaling to more implicit ambiguities in user queries or with clarification references in external knowledge bases would be critical.

\section*{Acknowledgments}
We are grateful to the members of Amelia Science, R\&D, IPsoft, Bangalore for their invaluable suggestions.

\bibliography{anthology,emnlp2020}
\bibliographystyle{acl_natbib}
\end{document}